\ifdefined\pdfminorversion
\fi
\documentclass[runningheads]{llncs}

\usepackage{amsmath,amssymb,amsfonts}
\usepackage{booktabs}
\usepackage{cite}
\usepackage{float}
\usepackage{graphicx}
\usepackage{hyperref}

\graphicspath{{figs/}}
\hypersetup{
  pdftitle={Symbol and Footprint Database for Electronic Components by Agentic Recognition and Generation},
  pdfauthor={Yichen Shi, Yuzhi Liu, Zhuofu Tao, Li Huang, Yuhao Gao, Ting-Jung Lin, and Lei He},
  hidelinks
}

\begin{document}

\title{Symbol and Footprint Database for Electronic Components by Agentic Recognition and Generation}
\titlerunning{Symbol and Footprint Database for Electronic Components}

\author{%
  Yichen Shi\inst{1,2}\textsuperscript{*} \and
  Yuzhi Liu\inst{3}\textsuperscript{*} \and
  Zhuofu Tao\inst{3} \and
  Li Huang\inst{1,2} \and
  Yuhao Gao\inst{3} \and
  Ting-Jung Lin\inst{1}\textsuperscript{**} \and
  Lei He\inst{1}\textsuperscript{**}%
}
\authorrunning{Y. Shi et al.}

\institute{%
  Ningbo Institute of Digital Twin, Eastern Institute of Technology, Ningbo, China\\
  \email{tlin@idt.eitech.edu.cn, lhe@eitech.edu.cn}
  \and
  Shanghai Jiao Tong University, Shanghai, China\\
  \email{shiyichen@sjtu.edu.cn}
  \and
  BTD Technology, Ningbo, China\\
  \email{yuzhi.liu@btd.tech}
}

\maketitle

\renewcommand{\thefootnote}{\fnsymbol{footnote}}
\footnotetext[1]{Equal contribution.}
\footnotetext[2]{Corresponding authors.}
\renewcommand{\thefootnote}{\arabic{footnote}}

\begin{abstract}
A rich and recognizable component library is a cornerstone of printed circuit board (PCB) design and generation. Traditionally, engineers manually create symbols and footprints for PCB schematics, a process that is time-consuming and error-prone. Leveraging multimodal large language models (MLLMs), we develop SFgen, an agentic workflow for recognizing and generating electronic-component symbols and footprints. SFgen achieves 86\% accuracy for symbol generation and 80\% accuracy for footprint generation. Using SFgen, we construct SFnet, a database containing symbols and footprints for 1,000 components, which provides a foundation for automated PCB design generation.

\end{abstract}

\section{Introduction}
Component libraries are crucial in printed circuit board (PCB) design, offering designers a predefined set of electronic components for both PCB schematics and layouts. These libraries provide a comprehensive set of resources, including datasheets, schematic symbols, 2D footprints, and 3D models.

Traditionally, companies manually create symbols and footprints based on datasheets to build component libraries. This manual process is labor-intensive and error-prone. 
Additionally, the varied formats of datasheets from different vendors complicate the implementation of a standardized design process. The differing requirements of various electronic design automation (EDA) software further add to the challenge, making it difficult to reuse library resources across design platforms and thus increasing development costs. With the continuous influx of new components into the market each year, a more efficient and automatic method of constructing electronic component libraries is essential to keep pace with technological advancements and market demands.

Datasheet formats are heterogeneous, and large-scale datasets are scarce. These factors make it difficult for traditional artificial intelligence (AI) methods to capture component design details. Multimodal large language models (MLLMs)~\cite{gpt,gemini,llama} offer a promising way to extract essential information from datasheets and generate design files. Prompt-engineering methods~\cite{cot,vcot,som,lin2024layoutprompter,wang2024qwen2} further help MLLMs solve these tasks without retraining. As electronic systems grow more complex, AI-assisted PCB design also offers an opportunity to shorten design cycles while maintaining design quality. Automatic component-library construction is a necessary foundation for MLLMs to understand and create PCB designs.

\begin{figure}
    \centering
    \includegraphics[width=0.8\textwidth]{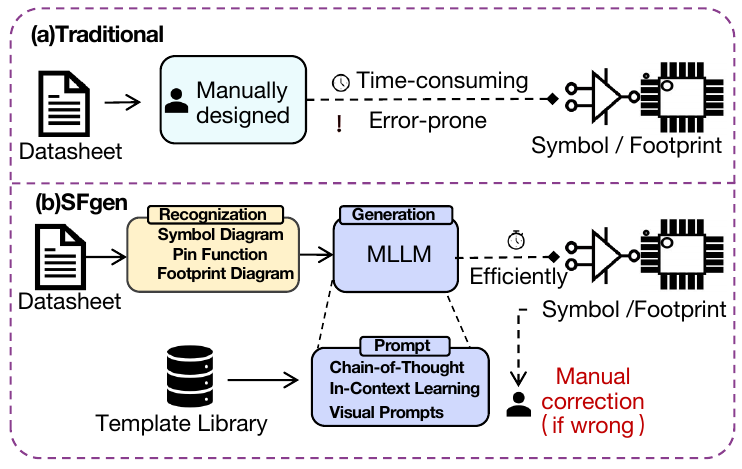}
    \caption{Comparison of the traditional manual design process and SFgen.}
    \label{fig:pcb_workflow}
\end{figure}

This paper presents SFgen, an automatic pipeline for constructing electronic component libraries. As summarized in Fig.~\ref{fig:pcb_workflow}, SFgen uses an MLLM to parse datasheets and generate symbols and footprints for electronic components. Based on SFgen, we create SFnet, a dataset containing more than 1,000 electronic components.

To assess the performance of generated designs, we introduce both qualitative and quantitative evaluation metrics. The qualitative evaluation focuses on the aesthetic appeal, and the quantitative metrics measure the accuracy of pin/shape counts and deviations in geometric dimensions. In addition, we construct a benchmark dataset and evaluation protocol, consisting of 100 symbols and footprints, each accompanied by its corresponding datasheet, to facilitate rigorous testing and comparison.


To the best of our knowledge, SFgen is the first published study of MLLM-driven agentic recognition and generation of PCB component designs. The proposed method and the resulting SFnet dataset lay a solid foundation for the automatic generation of PCB designs in the future. Our main contributions are summarized as follows.
\begin{enumerate}
\item We propose SFgen, an MLLM-based agentic solution for symbol and footprint generation of PCB components. 
\item We provide a benchmark testset consisting of datasheets and their corresponding symbol and footprint files along with the evaluation protocol. The benchmark facilitates rigorous testing and comparison of different methods.
\item Based on SFgen, we construct SFnet, a database encompassing more than 1,000 symbols and footprints of electronic components. As we know, this is the first large-scale dataset for automatic electronic component generation.


\end{enumerate}

\section{SFgen Algorithms}

This section provides the details of each step of SFgen.

\subsection{Problem Definition}
We define the automatic symbol and footprint generation as a conditional text generation task, as most EDA software stores these content in text formats. Inputs to the pipeline are datasheets containing pin configurations, pin function descriptions, and footprint details, etc. The output symbols and footprints should follow specific EDA software formats. 

A component symbol refers to the graphical symbol used in schematic diagrams. It encompasses critical information for maintaining the precision and integrity of the circuit design, including (1) pin function: the specific role of each pin within a circuit, and (2) pin configuration: the layout and physical arrangement of pins on a component or chip. A complete symbol should include correct name, location, and type for each pin.

The footprint file is foundational for creating the layout, ensuring the correct installation of an electronic component. The essential information includes (1) pad sizes, (2) pad placements, and (3) component body outline (the actual space occupied by the component, which helps avoid physical conflicts between components in layout). A complete footprint file should include the perimeter coordinates for each pad polygon.

\subsection{Algorithm Overview}
Considering the conditional text generation task, we denote the dataset $\mathcal{D} = \{ (x_j, y_j) \}_{j=1}^{M}$, where $(x_j, y_j)$ is the $j-$th sample of the dataset, an (input constraint, output text) pair, while $M$ is the total number of samples. 

\begin{figure}
    \centering
    \includegraphics[width=0.9\textwidth]{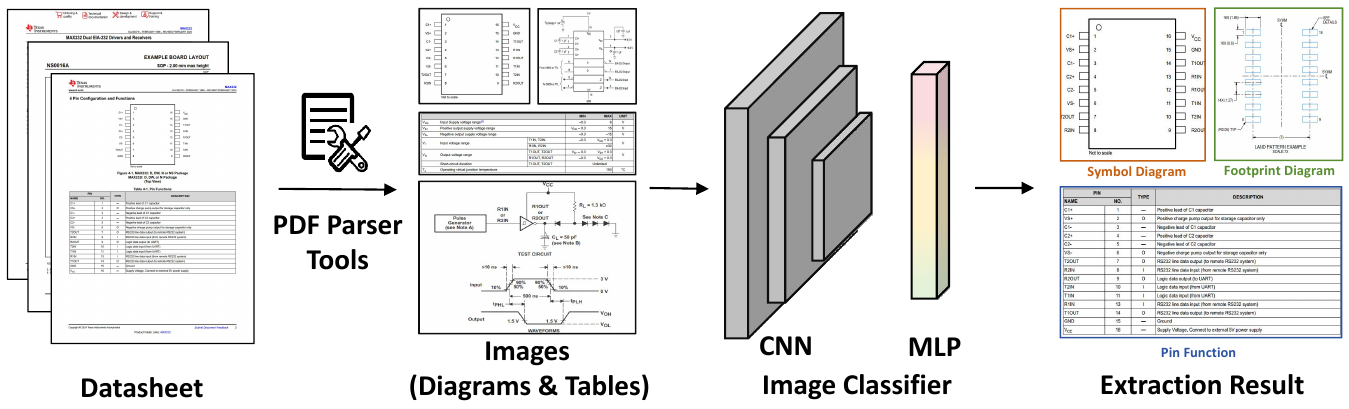}
    \caption{Information extraction from component datasheets.}
    \label{fig:datasheet_extraction}
\end{figure}

SFgen starts by extracting information from datasheets, as illustrated in Fig.~\ref{fig:datasheet_extraction}. Based on the open-source PDF parsing tool~\cite{wang2024mineruopensourcesolutionprecise}, the pipeline extracts images from datasheets and classifies them to obtain key constraints for component construction. These images include symbol diagrams, footprint diagrams, and pin-function tables. The pipeline also summarizes critical information such as the component manufacturer, device type, package format, and pin count. This information is used to retrieve the best-matching template from the template library for in-context learning (ICL).

As summarized in Fig.~\ref{fig:pcb_workflow}(b), $k$ samples $\{ (x_j, y_j) \}_{j=1}^{k}$ are provided to SFgen as ICL templates, where $x_j$ and $y_j$ are the constraint and target file, respectively. A test directive is based on the constraint $x_{\mathrm{test}}$, with the goal of generating a component-library file $y_{\mathrm{test}}$. The preamble $R$ specifies the task, such as ``symbol generation'' or ``footprint generation.'' The final prompt and output are formulated as follows:

\begin{equation}
    P = \left[ R; x_{1}; y_{1}; \ldots; x_{k}; y_{k}; x_{\mathrm{test}} \right];
    \qquad y = \operatorname{MLLM}(P).
    \label{eq:sfgen_prompt}
\end{equation}

\subsection{Symbol Generation}

\begin{figure*}[t]
    \centering
    \includegraphics[width=\textwidth]{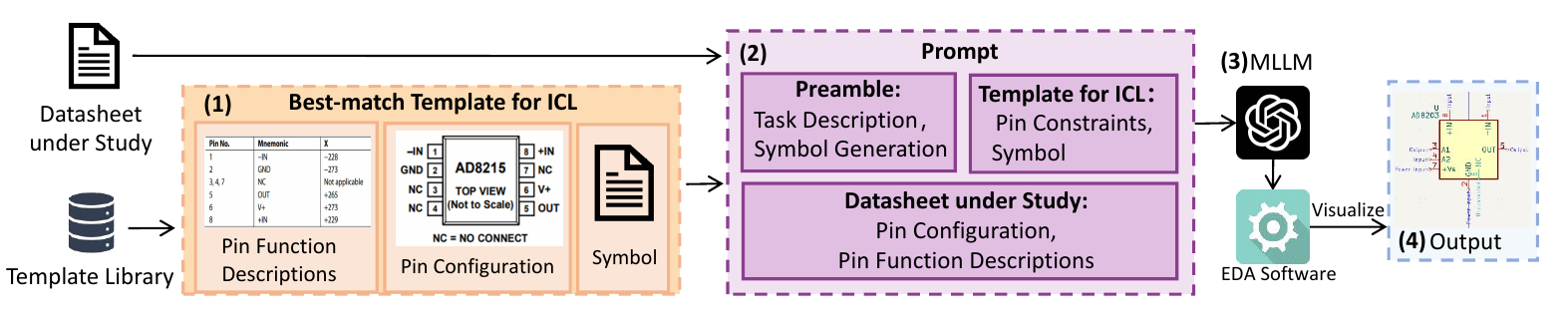}
    \caption{Symbol-generation flow: (1) select the best-matching ICL cases from the template library, (2) construct a prompt with multimodal information from the target datasheet, (3) generate a symbol file with the MLLM, and (4) visualize and compare the result against ground truth (GT).}
    \label{fig:symbol_pipeline}
\end{figure*}

Figure~\ref{fig:symbol_pipeline} illustrates the symbol-generation process employed by SFgen. The prompt integrates input constraints, ICL examples, task descriptions, and other relevant information. The MLLM then processes the inputs and provides the logical and spatial parameters needed to generate constraint-compliant symbol files. The few-shot ICL examples contain pin-function descriptions, configurations, and corresponding symbol files; they demonstrate the mapping between inputs and outputs and guide the MLLM to follow the required file format.

The improved accuracy ensures the correct number of pins and pin naming. Furthermore, SFgen learns to create legible symbol files by considering aspects such as polygon dimensions, the interconnections between lines and polygons, spacing between lines, and text formatting. This multifaceted learning approach equips SFgen to accommodate various file formats. These files can be visualized on dedicated EDA software, which also allows for further manual modifications, thereby reducing the manual effort required in the symbol generation process.

\subsection{Footprint Generation}
\begin{figure*}[t]
    \centering
    \includegraphics[width=\textwidth]{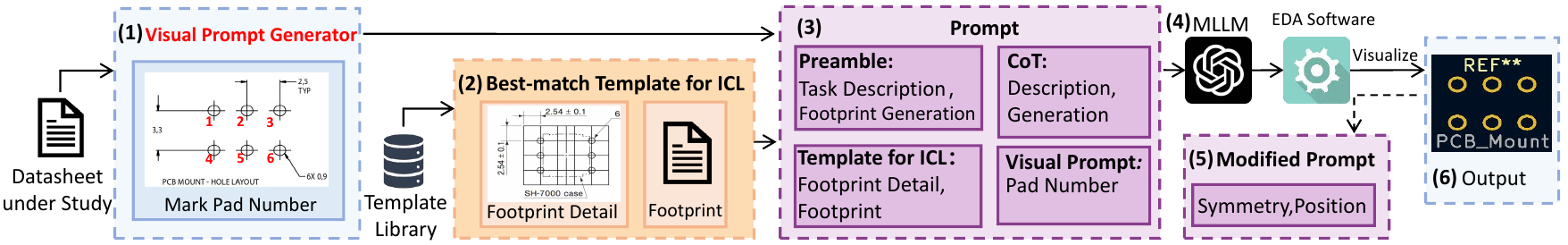}
    \caption{Footprint-generation flow: (1) attach visual prompts to datasheet images, (2) select the best-matching ICL cases from the template library, (3) construct the full prompt, (4) generate a footprint with the MLLM, (5) incrementally improve the footprint through prompt modification, and (6) compare the result against GT.}
    \label{fig:footprint_pipeline}
\end{figure*}

The footprint-generation flow is summarized in Fig.~\ref{fig:footprint_pipeline}. First, footprint information is extracted from the input datasheet, including the size and position of each pad. As in symbol generation, SFgen uses few-shot ICL examples to help the MLLM understand constraints on pad locations and sizes. SFgen can generate different file formats by selecting the corresponding template format.

\begin{figure}[t]
    \centering
    \includegraphics[width=0.5\textwidth]{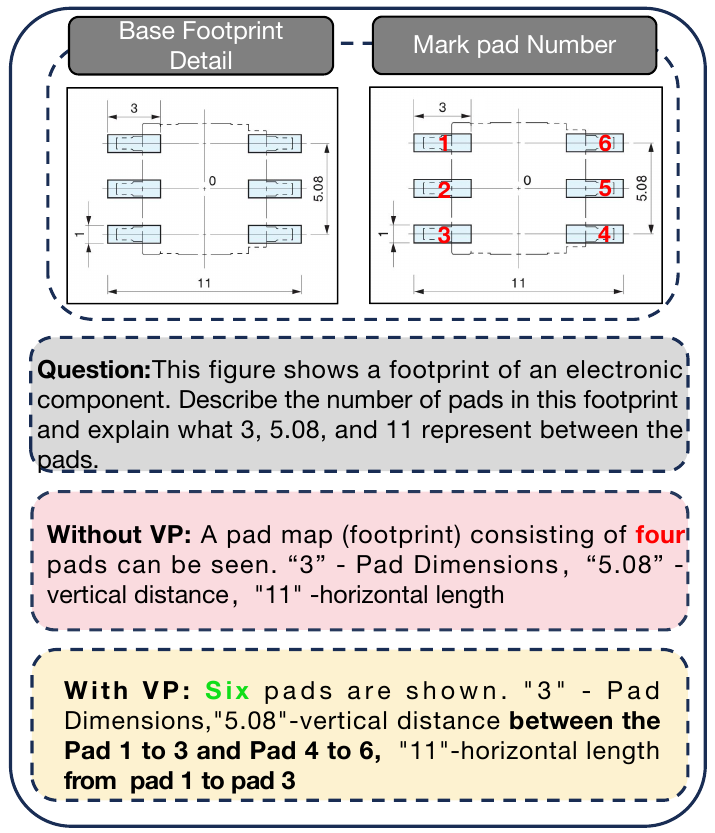}
    \caption{Illustration of visual prompting (VP): without VP, the MLLM incorrectly identifies four pads; with VP, it correctly identifies six pads.}
    \label{fig:visual_prompt}
\end{figure}

Unlike symbol generation, footprint generation uses visual prompting (VP) to improve SFgen's understanding of pad dimensions and positions. Figure~\ref{fig:visual_prompt} illustrates how the VP generator annotates each pad with a numerical identifier. These identifiers help the MLLM recognize the number, locations, and dimensions of the pads. We also introduce chain-of-thought (CoT) reasoning with visual information in the text prompt. The prompt asks the model to describe the pads using their assigned indices, guiding it to process the input more systematically.

Prompts for footprint generation integrate the constraints, templates, and add-on visual information. SFgen responds to the input by generating a preliminary footprint file that can be used by EDA software for visualization. However, the preliminary outputs may contain errors, particularly in the symmetry and positions of pads. Hence, we introduce the \textit{modified prompt}, which checks footprint and corrects errors step by step. Specifically, the visual images are re-imported into SFgen, and footprints are further optimized to ensure that the final outputs meet the design requirements.

\section{Evaluation Protocol}

\subsection{Benchmark Construction}

\begin{figure}
    \centering
    \includegraphics[width=0.8\textwidth]{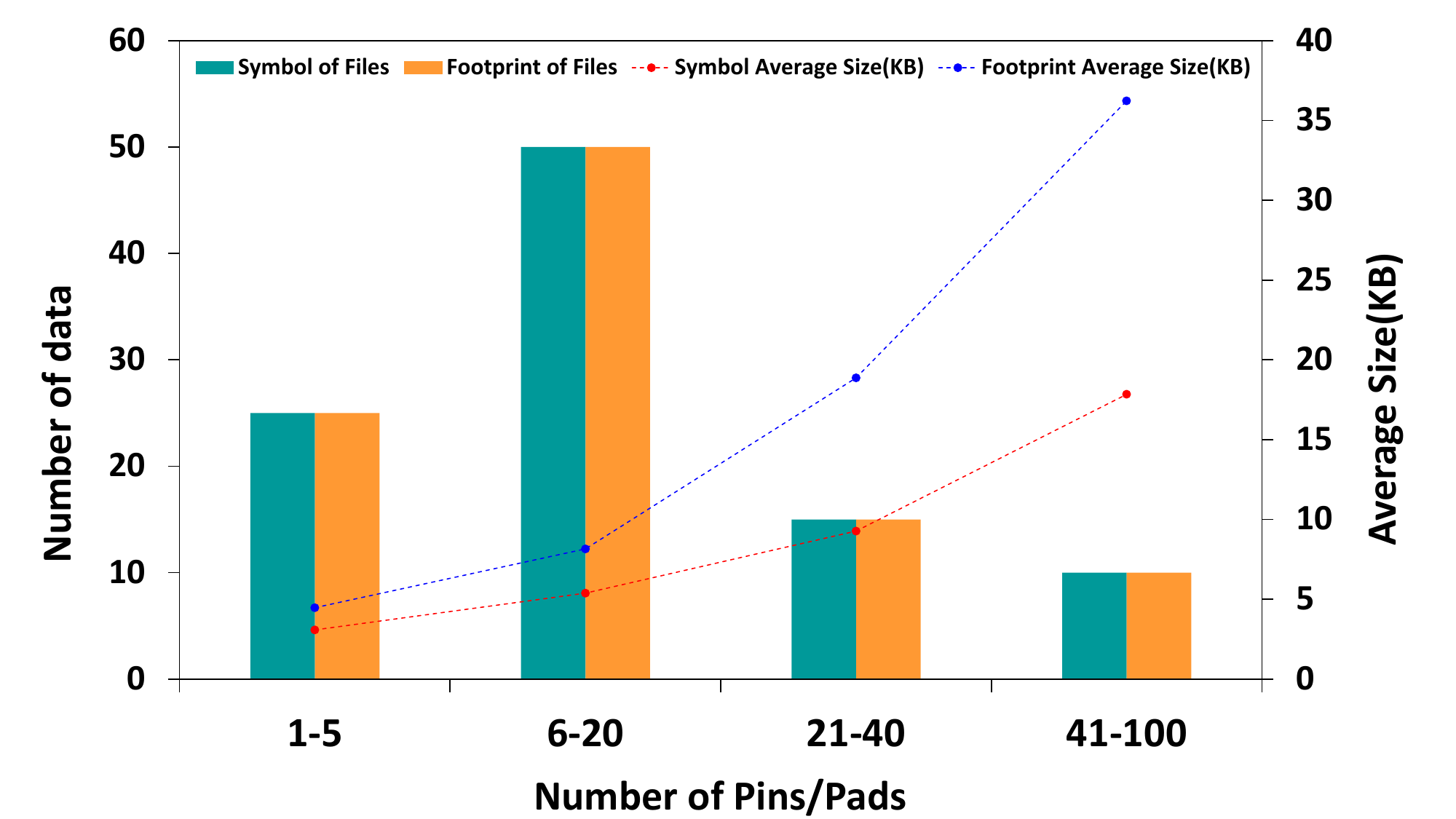}
    \caption{Distribution of components in the evaluation benchmark.}
    \label{fig:benchmark_distribution}
\end{figure}

Electronic components vary substantially in functionality and package configuration, directly affecting the complexity of symbol and footprint generation. To evaluate the reliability and compatibility of SFgen, we select a representative dataset based on pin count, device type, and the availability of both symbol and footprint data. Table~\ref{tab:benchmark} summarizes the benchmark, and Fig.~\ref{fig:benchmark_distribution} presents its distribution.

\begin{table*}[h] 
\centering
\caption{Evaluation benchmark.}
\label{tab:benchmark}
\renewcommand{\arraystretch}{1}
\resizebox{1\textwidth}{!}{
\begin{tabular}{p{2cm} p{3cm} p{3cm} p{5cm} p{2cm}} 
\toprule
\multicolumn{5}{c}{\textbf{Symbol and Footprint Generation}} \\ 
\midrule
\textbf{Group Type} & \textbf{Complexity Level} & \textbf{Pin-Number } & \textbf{Components Type} & \textbf{Quantity} \\ 
\midrule
Basic  & Simple & 1-5 & R, L, C, D, Fuse, LED, Switch & 25 \\ 
Standard  & Moderate & 6-20 & IC, Sensor, Transistor & 50 \\ 
Complex  & Difficult & 21-40 & IC, Module & 15 \\ 
High-Density  & Extremely Difficult & 41-100 & IC & 10 \\ 
\bottomrule
\end{tabular}
}
\end{table*}

\subsection{Evaluation Metrics}

To evaluate generated results, we first consider the counts and types of generated pins and pads. Any discrepancy in these quantities is treated as a prediction error. Equations~\eqref{eq:acc_n} and~\eqref{eq:acc_t} define $\mathrm{ACC}_{N}$ and $\mathrm{ACC}_{T}$, respectively. The former is the fraction of components with the correct number of pins or pads, while the latter compares correctly typed pins or pads with the total number. We compute $\mathrm{ACC}_{T}$ only when the generated pin or pad count is correct; otherwise, the entire component is considered incorrect.

\begin{equation}
    \mathrm{ACC}_{N} = \frac{\text{Components with correct pin/pad counts}}{\text{Total components}}.
    \label{eq:acc_n}
\end{equation}

\begin{equation}
    \mathrm{ACC}_{T} = \frac{\text{Correctly typed pins/pads}}{\text{Total pins/pads}}.
    \label{eq:acc_t}
\end{equation}

We assess footprint size and position only when pad types and counts are correct. The sum of normalized area differences measures the discrepancy between generated and ground-truth pad sizes. Similarly, the normalized coordinate differences measure positional accuracy. Equations~\eqref{eq:dif_a} and~\eqref{eq:dif_p} define $\mathrm{Dif}_{A}$ and $\mathrm{Dif}_{P}$, respectively.

\begin{equation}
    \mathrm{Dif}_{A} = \sum_{i=1}^{N} \frac{|\mathrm{Area}_{\mathrm{gen},i} - \mathrm{Area}_{\mathrm{GT},i}|}{\mathrm{Area}_{\mathrm{GT},i}}.
    \label{eq:dif_a}
\end{equation}

\begin{equation}
   \mathrm{Dif}_{P} = \sum_{i=1}^{N} \frac{|\mathrm{Pos}_{\mathrm{gen},i} - \mathrm{Pos}_{\mathrm{GT},i}|}{\sqrt{\mathrm{Area}_{\mathrm{GT},i}}}.
   \label{eq:dif_p}
\end{equation}

In these equations, $\mathrm{Area}$ and $\mathrm{Pos}$ denote the area and position of a pad, $i$ indexes pads, and the subscripts $\mathrm{gen}$ and $\mathrm{GT}$ denote the SFgen output and ground truth, respectively.



Other than quantitative evaluations, we design \textit{mean opinion score for symbol (MOS-S)} and \textit{mean opinion score for footprint (MOS-F)} based on designer expertise to thoroughly assess the design quality. Assessments are conducted from four dimensions: accuracy, readability, usability, and aesthetics. Evaluators rate the overall quality of the generated symbols and footprints on a scale ranging from 1 to 5, with higher scores denoting superior quality. As a subjective evaluation method, the scores evaluate how the generated design style  conforms to the existing ones in practical applications. Table \ref{tab:mos} outlines the specific scoring criteria.

\begin{table}[h] 
\centering
\caption{MOS Evaluation Criteria}
\label{tab:mos}
\resizebox{0.8\textwidth}{!}{
\begin{tabular}{l c l} 
\toprule
Quality Level & MOS &    Evaluation Criteria \\
\midrule
Excellent & 5 & Excellent quality; no adjustments. \\
Good & 4 & Good quality; only minor adjustments. \\
Average & 3 & Average quality; some adjustments. \\
Poor & 2 & Poor quality; numerous adjustments. \\
Inferior & 1 & Inferior quality; significant adjustments. \\
\bottomrule
\end{tabular}
}
\end{table}



\section{Experiments}

\subsection{Experiment Setting}

We implement SFgen using GPT-4 through API calls and visualize the generated files with KiCad 8.0.5. The full prompts are provided in Figs.~\ref{fig:prompt_symbol}, \ref{fig:prompt_modification}, and~\ref{fig:prompt_footprint} in the appendix.

\subsection{Symbol Generation Results}

\begin{table}[t]
  \centering
  \caption{Evaluation results for generated symbols.}
  \label{tab:symbol_results}
  \resizebox{0.6\textwidth}{!}{%
    \begin{tabular}{lccc}
      \toprule
      Group & $\mathrm{ACC}_{N}\,\uparrow$ & $\mathrm{ACC}_{T}\,\uparrow$ & MOS-S \\
      \midrule
      Basic        & 96.00\% & 96.36\% & 5 \\
      Standard     & 86.00\% & 83.46\% & 4 \\
      Complex      & 73.33\% & 84.80\% & 3 \\
      High-Density &  0.00\% & 49.32\% & 1 \\
      \bottomrule
    \end{tabular}%
  }
\end{table}

The two automatic metrics, $\mathrm{ACC}_{N}$ and $\mathrm{ACC}_{T}$, together with MOS-S, measure the quality of the generated symbols. Table~\ref{tab:symbol_results} presents SFgen results across the four benchmark groups. In the Basic group, 96\% of the generated symbols have the correct number of pins. Performance decreases as complexity increases. Even in the High-Density group, where no output has the exact pin count, SFgen achieves a pin-type accuracy of 49.32\%.

\begin{figure*}
    \centering
    \includegraphics[width=\textwidth]{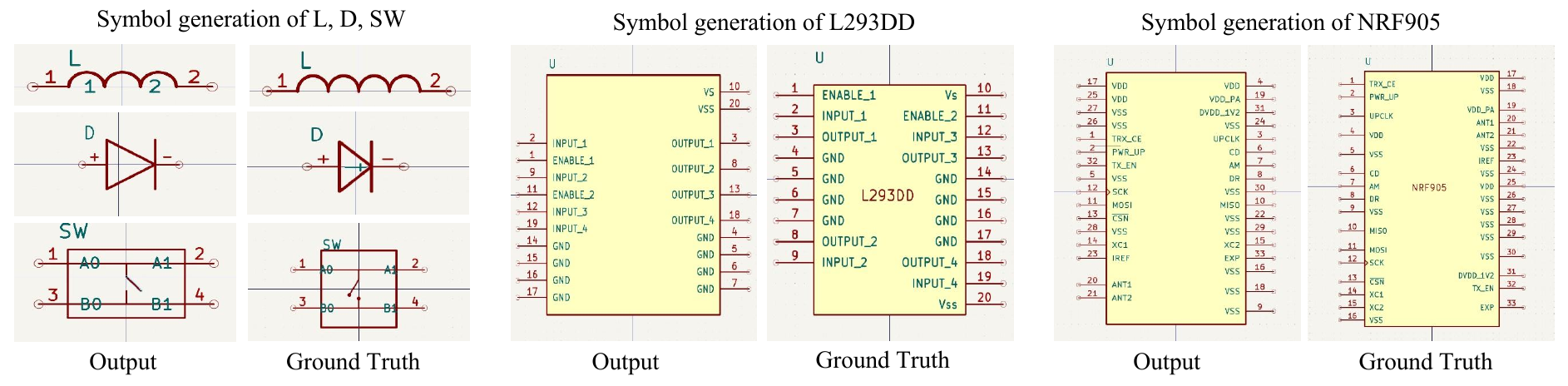}
    \caption{Visualization results of generated symbols}
    \label{fig:symbol_results}
\end{figure*}

Figure~\ref{fig:symbol_results} compares generated symbols with ground truth (GT). The outputs are visually consistent with GT and capture functional layouts, including pin labels and positions, while remaining readable.



\subsection{Footprint Generation Results}
The four automatic metrics, $\mathrm{ACC}_{N}$, $\mathrm{ACC}_{T}$, $\mathrm{Dif}_{A}$, and $\mathrm{Dif}_{P}$, together with MOS-F, measure the quality of generated footprints. Table~\ref{tab:footprint_results} summarizes the results. SFgen achieves $\mathrm{ACC}_{N}$ values of 96\% and 80\% in the Basic and Standard groups, respectively, while $\mathrm{Dif}_{A}$ remains below 10\% in both groups. Performance declines as footprint complexity increases, with a sharp drop in the High-Density group because of the MLLM's limited ability to recognize complex images.

\begin{table}[t]
  \centering
  \caption{Evaluation results for generated footprints.}
  \label{tab:footprint_results}
  \resizebox{0.8\textwidth}{!}{%
    \begin{tabular}{lccccc}
      \toprule
      Group & $\mathrm{ACC}_{N}\,\uparrow$ & $\mathrm{ACC}_{T}\,\uparrow$ & $\mathrm{Dif}_{A}\,\downarrow$ & $\mathrm{Dif}_{P}\,\downarrow$ & MOS-F \\
      \midrule
      Basic        & 96.00\% & 94.54\% &  4.26\% &  8.23\% & 5 \\
      Standard     & 80.00\% & 89.80\% &  8.76\% & 15.30\% & 4 \\
      Complex      & 46.67\% & 65.59\% & 11.18\% & 28.87\% & 2 \\
      High-Density &  0.00\% & 35.40\% & 19.32\% & 42.50\% & 1 \\
      \bottomrule
    \end{tabular}%
  }
\end{table}

\begin{figure*}[t]
    \centering
    \includegraphics[width=\textwidth]{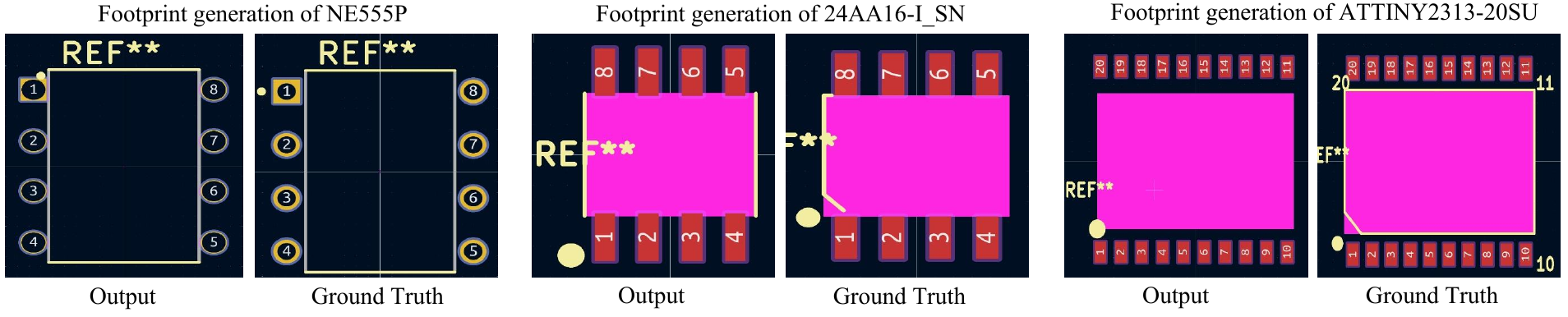}
    \caption{Visualization results of generated footprints}
    \label{fig:footprint_results}
\end{figure*}

Figure~\ref{fig:footprint_results} shows representative footprint visualizations. The generated footprints capture pad shapes, sizes, and spacing. Because the generated files are text-based and editable, designers can correct remaining discrepancies in an EDA platform.

\subsection{Ablation Study for Footprint Generation}

 \begin{figure*}
    \centering
    \includegraphics[width=\textwidth]{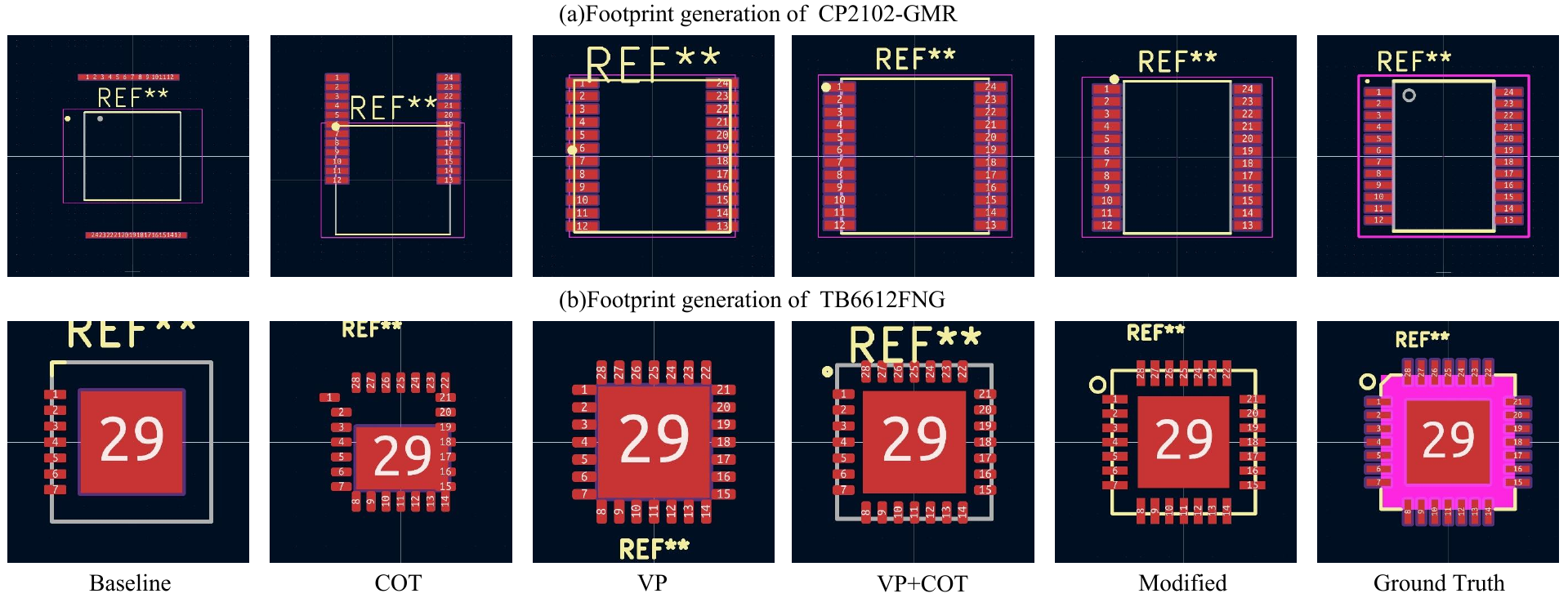}
    \caption{Ablation study of footprint generation}
    \label{fig:ablation_results}
\end{figure*}

We conduct a limited ablation study of VP, CoT, and prompt modification for footprint generation. As shown in Fig.~\ref{fig:ablation_results}, CoT alone produces suboptimal results for complex components. Annotating each pad through VP and combining the annotations with CoT improves pad-position accuracy. A subsequent prompt-modification step further refines the generated footprint.

\section{SFnet Database}

Based on SFgen, we develop SFnet, which contains 1,000 electronic-component datasheets and their corresponding symbols and footprints. Table~\ref{tab:sfnet_distribution} summarizes the component categories; integrated circuits (ICs) constitute the majority. To our knowledge, SFnet is the first large-scale dataset of electronic-component symbols and footprints.



\begin{table}[t]
    \centering
    \caption{Category distribution of the SFnet dataset.}
    \label{tab:sfnet_distribution}
    \begin{tabular}{lr@{\hspace{1.5cm}}lr}
    \toprule
    \textbf{Component Type} & \textbf{Quantity} & \textbf{Component Type} & \textbf{Quantity} \\ 
    \midrule
    IC & 722 & Resistor & 165 \\ 
    Capacitor & 37 & Sensor & 19 \\ 
    Transistor & 15 & Diode & 11 \\ 
    Inductor & 6 & LED & 6 \\ 
    Fuse & 5 & Module & 4 \\ 
    MOS & 3 & Crystal Oscillator & 3 \\ 
    Display & 1 & ESD & 1 \\ 
    Antenna & 1 & Relay & 1 \\ 
    \bottomrule
    \end{tabular}  
\end{table}

\section{Conclusions and Future Work}

This paper presents SFgen, an MLLM-driven pipeline that extracts component information from datasheets and generates symbol and footprint files. SFgen combines in-context learning (ICL), visual prompting (VP), and chain-of-thought (CoT) reasoning to capture design details. It generates a component in 3--5 minutes and achieves 86\% symbol-generation accuracy and 80\% footprint-generation accuracy. Based on SFgen, we construct SFnet, a database containing datasheets, symbols, and footprints for 1,000 electronic components.

Future work will extend SFgen to a broader range of components and release SFnet to support further research. We also plan to investigate how MLLMs can use component libraries to understand PCB schematics and assist automated PCB design.

\bibliographystyle{splncs04}
\bibliography{refs}

\clearpage
\appendix
\section{Prompt Details}

\begin{figure}[htbp]
  \centering
  \includegraphics[width=0.65\textwidth]{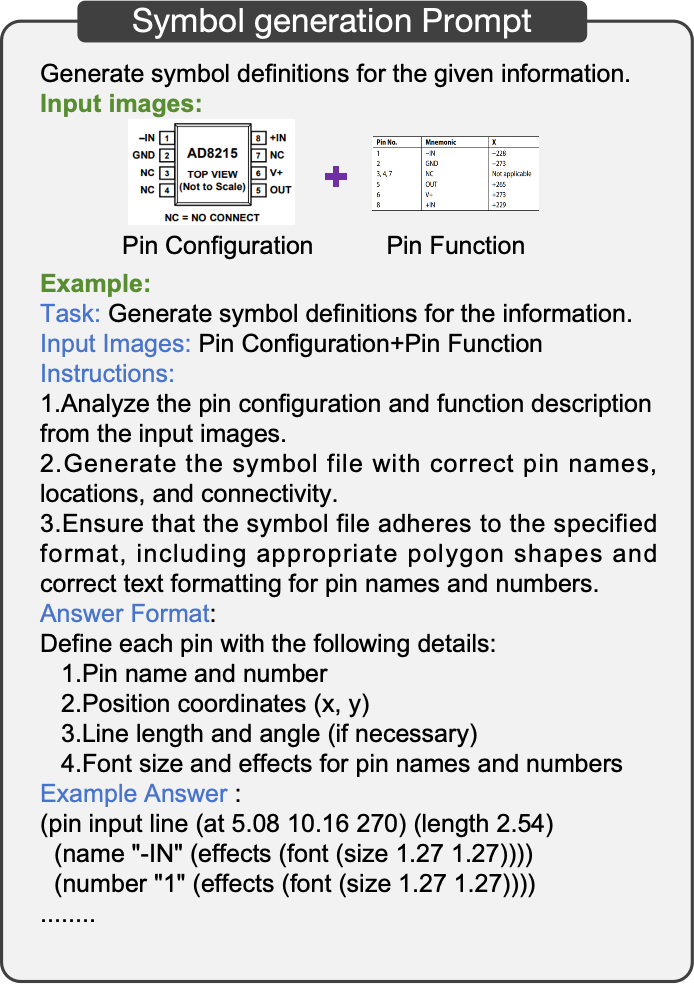}
  \caption{Design prompt for symbol generation.}
  \label{fig:prompt_symbol}
\end{figure}

\begin{figure}[htbp]
  \centering
  \includegraphics[width=0.85\textwidth]{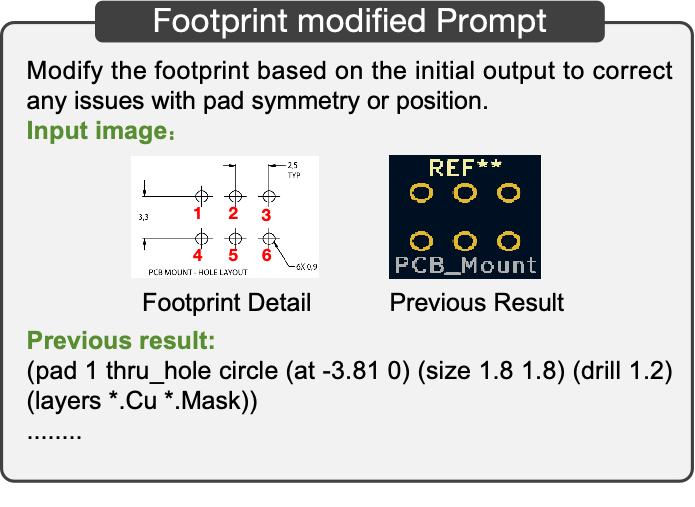}
  \caption{Modification prompt for footprint generation.}
  \label{fig:prompt_modification}
\end{figure}

\begin{figure}[htbp]
  \centering
  \includegraphics[width=0.65\textwidth]{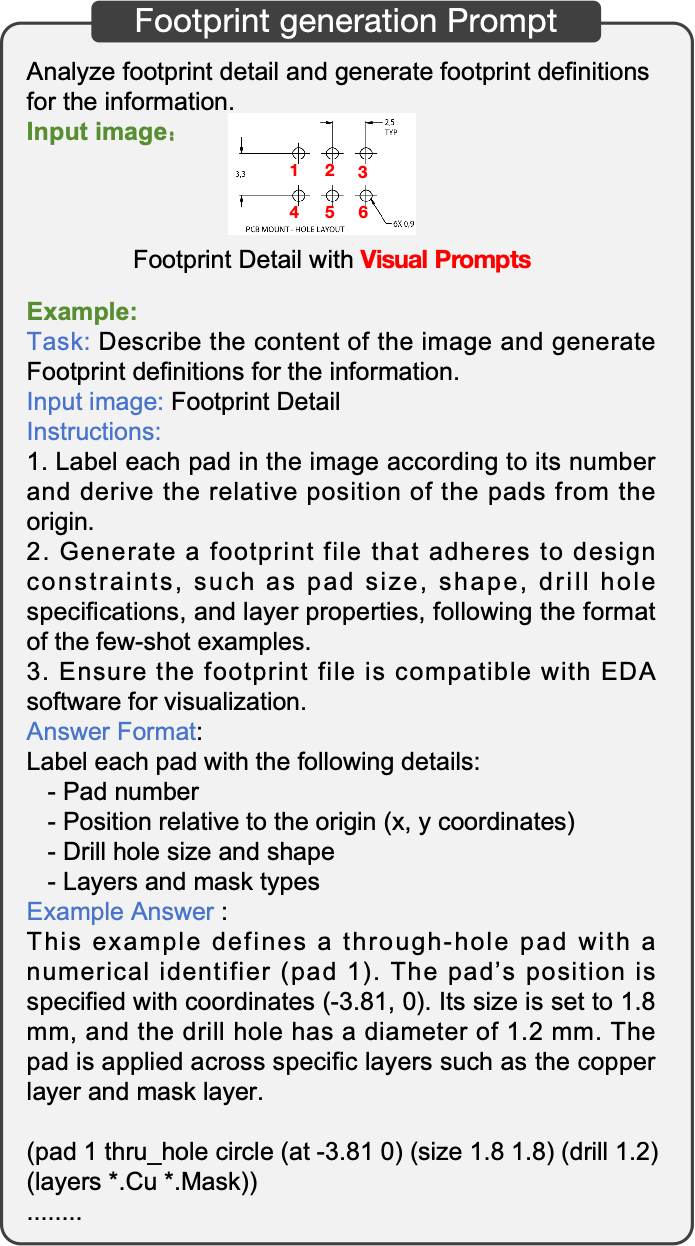}
  \caption{Design prompt for footprint generation.}
  \label{fig:prompt_footprint}
\end{figure}

\end{document}